\documentclass[a4paper,fleqn]{cas-dc}

\usepackage[numbers,square,sort&compress]{natbib}
\usepackage{threeparttable}
\usepackage{soul}
\usepackage{xcolor}
\usepackage{booktabs}
\usepackage{threeparttable}
\usepackage{graphicx}
\definecolor{hlBcolor}{RGB}{173,216,230} 

\def\tsc#1{\csdef{#1}{\textsc{\lowercase{#1}}\xspace}}
\tsc{WGM}
\tsc{QE}
\tsc{EP}
\tsc{PMS}
\tsc{BEC}
\tsc{DE}
\ExplSyntaxOn
\cs_gset:Npn \__first_footerline:
{
  \group_begin:
  \small
  \sffamily
  \__short_authors:
  \group_end:
}
\ExplSyntaxOff
\begin{document}
\let\WriteBookmarks\relax
\def\floatpagepagefraction{1}
\def\textpagefraction{.001}
\shorttitle{HUMAN-TCI: Torso-Centered Interaction for Text-to-Motion Retrieval}
\shortauthors{}

\title{
\texorpdfstring{
\includegraphics[height=1em]{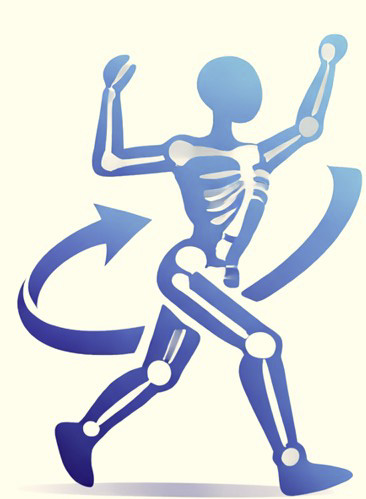}\hspace{0.1em}
HU{\textbf{\color{blue}M}}AN-TCI: 
{\color{blue}H}ierarchical {\color{blue}M}ulti-Stream 
{\color{blue}M}otion-{\color{blue}A}ware 
{\color{blue}N}etwork with 
{\color{blue}T}orso-{\color{blue}C}entered 
{\color{blue}I}nteraction for Text-to-Motion Retrieval
}
{
HU-MAN-TCI: Hierarchical Multi-Stream Motion-Aware Network with Torso-Centered Interaction for Text-to-Motion Retrieval
}
}

\author[1,2]{Muhammad Islam}[
    orcid=0000-0002-4845-4808]

\ead{muhammad.islam1@jcu.edu.au}
\ead[url]{https://www.jcu.edu.au/}

\credit{Conceptualization, Methodology, Software, Data Curation, Writing--Original Draft Preparation}


\author[1,2]{Euijoon Ahn}[
    orcid=0000-0001-7027-067X]

\ead{euijoon.ahn@jcu.edu.au}
\ead[url]{https://www.jcu.edu.au/}

\credit{Supervision, Writing--Review \& Editing}


\author[3]{Usman Naseem}[
    orcid=0000-0003-0191-7171]

\ead{usman.naseem@mq.edu.au}
\ead[url]{https://www.mq.edu.au/}

\credit{Supervision, Writing--Review \& Editing}


\author[1,2]{Tao Huang}[
    orcid=0000-0002-8098-8906]

\ead{tao.huang1@jcu.edu.au}
\ead[url]{https://www.jcu.edu.au/}

\credit{Supervision, Writing--Review \& Editing}


\affiliation[1]{organization={College of Science and Engineering, James Cook University},
                city={Cairns},
                postcode={4878},
                state={QLD},
                country={Australia}}

\affiliation[2]{organization={Centre for AI and Data Science Innovation, James Cook University},
                city={Cairns},
                postcode={4878},
                state={QLD},
                country={Australia}}

\affiliation[3]{organization={School of Computing, Macquarie University},
                city={Sydney},
                postcode={2113},
                state={NSW},
                country={Australia}}

\begin{abstract}
Accurate retrieval of human motions is a crucial first step in text-guided human motion modeling and synthesis, as it selects semantically relevant sequences from large datasets and provides grounded references for downstream tasks. Retrieving motions from natural language descriptions remains challenging because sentences can describe multiple actions, overlapping movements, and intricate dependencies between body parts. Existing methods often focus on simple, single-action descriptions and typically process body parts independently or by merely concatenating features, without explicitly modeling how torso movements influence other parts. In addition, their processing pipelines often rely on computationally heavy models, introducing considerable overhead, particularly when modeling longer or more complex motion sequences. This limits learning discriminative motion-pattern representations, reducing retrieval accuracy, interpretability, and efficiency in practical applications. To address these limitations, we propose HUMAN-TCI, a Hierarchical Multi-Stream Motion-Aware Network for text-guided human motion retrieval. HUMAN-TCI employs a three-stream architecture that separately models upper-body, lower-body, and torso motions while explicitly capturing their interactions, allowing torso-related movements to influence the positioning and dynamics of other body parts. By incorporating tailored torso attention,  our model effectively recognizes complex human motion patterns, captures fine-grained motion relationships and handles complex multi-action descriptions. Our framework supports retrieval for both simple, single-action sentences and long, compositional descriptions containing sequential or overlapping actions without relying on complex models. Evaluation on the KIT Motion-Language Dataset and HumanML3D demonstrates that HUMAN-TCI achieves better performance compared with prior conventional methods under the evaluated settings. Experimental results confirm that our approach retrieves semantically accurate, plausible, and interpretable motion sequences while remaining computationally efficient, making it suitable for large-scale, real-world applications.  \noindent\textbf{Project page (including code and visualization videos):}
\underline{\href{https://github.com/HUMAN-TCI/Hierarchical-Multi-Stream-Motion-Aware-Network-with-Torso-Centered-Interaction-}{GitHub HUMAN-TCI}}


\end{abstract}




\maketitle

\section{Introduction}
Retrieving human motion sequences from natural language descriptions is an essential capability for practical artificial intelligence (AI) systems. In real-world applications such as animation generation, robotics, virtual reality, and human-computer interaction, users often describe desired actions in natural language, and systems must translate these descriptions into accurate motion sequences \cite{islam2026multimodal,FineMLD}. Without reliable retrieval, it becomes difficult to locate semantically relevant motion data from large-scale motion repositories, limiting the ability to reuse, analyze, or synthesize meaningful motion patterns. However, this task is inherently challenging because natural language is ambiguous and compositional. A single sentence can describe multiple concurrent or sequential actions with variations in duration, intensity, and body-part involvement. For example, a description such as “a person is raising their arms while walking forward and turning their torso” involves coordinated upper-body, lower-body, and torso movements that must be jointly understood to retrieve the correct motion. Capturing such fine-grained and structured semantics remains a significant challenge for existing methods.

Existing text-to-motion retrieval methods typically embed the entire motion sequence and textual description into a single global feature vector, which often overlooks fine-grained spatial and temporal dependencies \cite{3}. Some approaches model body parts independently, treating the upper and lower body separately but simply concatenating torso features without capturing their influence on other joints \cite{h3,h5}. Other methods adopt hierarchical \cite{h2,h1} or multi-level learning strategies \cite{h4} to encode complex motions. Similarly, multi-instance learning approaches attempt to handle sequential or overlapping actions. However, none of these methods explicitly account for how torso movements affect other body parts or capture torso-related interactions, leaving an important gap in joint-level motion understanding and recognition. As a result, existing approaches often fail to distinguish co-occurring actions or capture nuanced correlations between language and joint-level motion, leading to semantically inaccurate or implausible retrieval results.


Another key challenge in motion retrieval is spatial grounding. Spatial grounding refers to the ability to align semantic concepts in text with specific spatial regions of the human body, such as mapping references like “arms,” “legs,” or “torso” to their corresponding joints and motion patterns. 
Human motion is inherently hierarchical, as the body forms an articulated kinematic structure with parent–child relationships between joints (e.g., torso, upper body, and lower body). While this structure governs how motion propagates across the body, coordinated interactions between these regions jointly define the overall motion semantics. However, many prior works ~\cite{3,3one} either treat the body as a single unit or rely on simplified upper- and lower-body representations, limiting their ability to model torso-centric interaction (TCI) and cross-part dependencies (CPD).

This limitation is evident in compositional descriptions involving multiple body parts. For example, in a sentence such as “a person is bending forward while stepping sideways,” accurate retrieval requires capturing both the lateral leg movement and the forward torso inclination. Existing approaches often fail to explicitly model these fine-grained dependencies, particularly when multiple coordinated actions are involved~\cite{2}, limiting their effectiveness in complex real-world scenarios.
To address these challenges, we propose HUMAN-TCI (Torso-Centered Interaction), a Hierarchical Multi-Stream Motion-Aware Network for text-guided human motion retrieval. Our model employs a three-stream architecture to separately encode upper-body, lower-body, and torso motions, and a hierarchical attention mechanism to model their interactions. This design better aligns textual descriptions with body-part-specific motion, improving retrieval for both simple and compositional actions. Prior sequential models~\cite{2,3} model motions using separate streams for different modalities and capture temporal dependencies sequentially. In contrast, our approach achieves more accurate, plausible, and interpretable retrieval results without requiring excessively complex architectures.


Our results indicate that incorporating hierarchical multi-stream modeling and torso-aware attention improves semantic alignment between text and motion representations. These gains are especially noticeable in queries involving complex, compositional actions with multiple interacting body parts.

\begin{figure*}[t]
    \centering
   \includegraphics[width=1\linewidth]{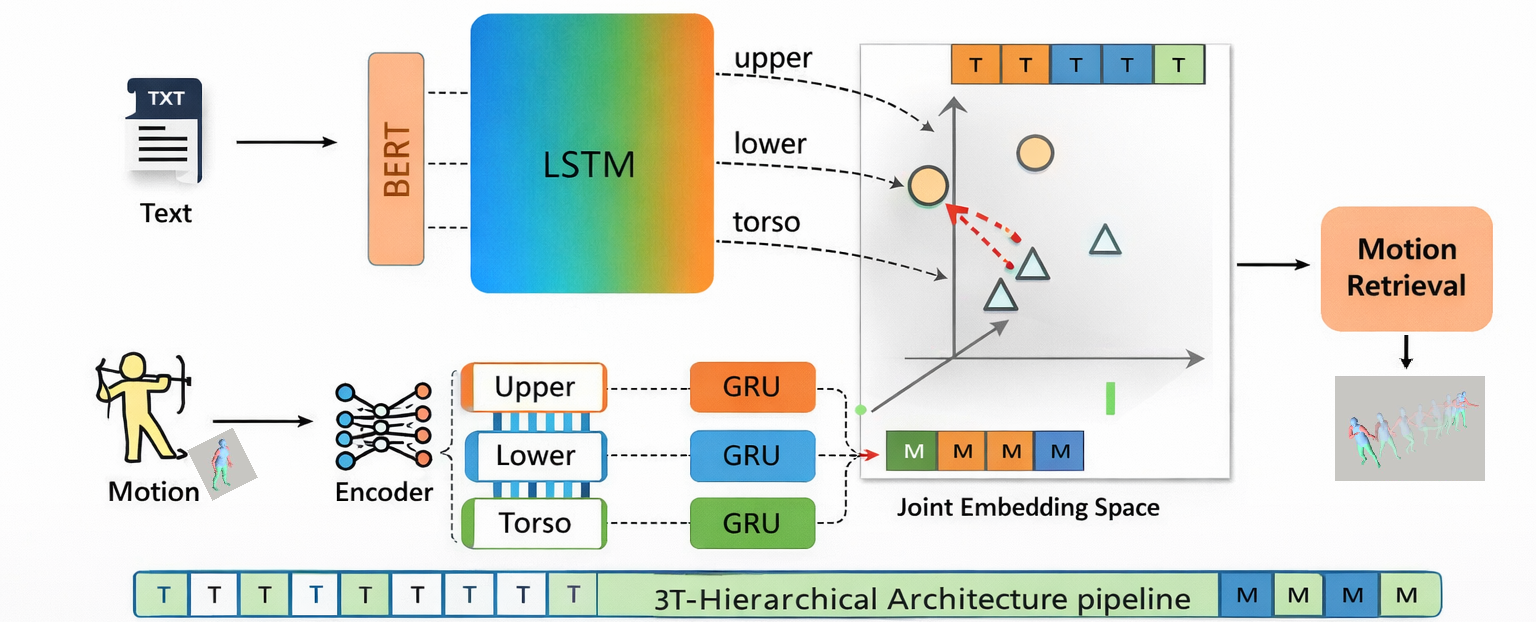}
    \caption{Overview of the proposed 3Tier-Hierarchical Architecture for text-to-motion retrieval. The framework learns hierarchical body-part-aware representations through dedicated upper-body, lower-body, and torso motion streams. Cross-stream dependencies are captured through torso-aware interaction (Figure~3), enabling fine-grained text--motion alignment in a shared embedding space. Triangles and circles represent text (T) and motion (M) embeddings, respectively, in the higher-dimensional embedding space. Details of the motion encoder are shown in Figure~2, while the text encoder is presented in Figure~4.}
    \label{fig:mainflow}
\end{figure*}

To the best of our knowledge, this work is among early retrieval frameworks to explicitly model torso-centered cross-stream interactions as shown in Figure~\ref{fig:mainflow}. By introducing a hierarchical multi-stream architecture with explicit torso-aware interaction modeling, HUMAN-TCI enables fine-grained spatial grounding between natural language descriptions and body-part motion dynamics, leading to more accurate and interpretable retrieval of complex multi-action motions. The main contributions of this work are summarized as follows:

\begin{itemize}
    \item We propose HUMAN-TCI, a hierarchical multi-stream architecture that decomposes human motion into upper-body, torso, and lower-body streams, enabling structured modeling of body-part dynamics for fine-grained text-to-motion retrieval.

\item We introduce an explicit TCI  mechanism that captures the influence of torso movements on both upper- and lower-body actions, allowing the model to better retrieve motions involving torso-driven dynamics such as bending, twisting, and turning.

\item Our approach improves the alignment between natural language descriptions and motion representations by explicitly modeling body-part-specific actions through a multi-stream design. This enables accurate, interpretable retrieval for both simple actions and complex compositional descriptions involving sequential and overlapping movements.

\item Extensive experiments on the KIT Motion-Language Dataset and HumanML3D demonstrate that HUMAN-TCI achieves consistent improvements in retrieval accuracy while maintaining a computationally efficient architecture compared to more complex retrieval frameworks.

\end{itemize}

\section{Related Work}

\subsection{Text-to-Motion Retrieval}

Text-to-motion retrieval (TMR) has recently emerged as an important task that aims to retrieve semantically relevant human motion sequences given a natural language description. Compared with traditional image–text or video–text retrieval tasks, TMR presents additional challenges due to the temporal dynamics of motion data \cite{temporal} and the compositional structure of human actions \cite{2,3,Gen}. Early work in this area formulates TMR by learning a shared embedding space that aligns motion sequences and textual descriptions using contrastive or triplet-based objectives \cite{yan2023cross}. 

Subsequent work establishes TMR as a standalone research problem by extending text-to-motion generation frameworks to retrieval settings and leveraging contrastive learning to align motion and language representations \cite{petrovich2023tmr}. Transformer-based architectures have also been adopted to improve motion representation learning, where motion sequences are encoded using transformer or vision transformer-style architectures and aligned with language encoders to enable cross-modal retrieval \cite{transformer, MoRAG}. 

More recent methods attempt to improve retrieval performance by introducing richer representations or additional supervision. For example, some approaches incorporate auxiliary modalities such as video to construct unified embedding spaces across text, video, and motion modalities \cite{trimodal}. Other works represent motion sequences as motion patches and apply transformer-based modeling to capture motion-language relationships \cite{transform1}. Event-level modeling has also been explored by decomposing textual descriptions into sequential sub-actions to better capture temporal correspondences between language and motion \cite{h2,h4,1}. 

Despite these advances, many existing TMR methods focus on aligning global motion representations with textual embeddings \cite{h5,segallgn, SemAllign}. Such global alignment often fails to capture the fine-grained spatial relationships between different body parts or the subtle semantics of complex multi-action descriptions, which are common in real-world motion-language datasets.

\subsection{Cross-Modal Semantic Alignment}

Cross-modal semantic alignment aims to learn a shared representation space where semantically related instances from different modalities are projected close together \cite{sem1, crossmodal,SAMR}. This paradigm has been widely explored in vision language tasks such as image–text and video–text retrieval. A common approach is to learn global embeddings for each modality and perform matching using similarity measures such as cosine similarity \cite{h2,h5}. While effective, global alignment strategies may overlook local correspondences that are important for capturing fine-grained semantic relationships.

To address this limitation, several methods introduce fine-grained alignment mechanisms that model interactions between visual regions and textual phrases or between video frames and words \cite{fineG1,sem1}. Researchers have also proposed hierarchical and multi-grained alignment strategies to capture semantic relationships at different levels of abstraction \cite{h1,2}. In addition, recent approaches leverage knowledge from large pre-trained models to enhance cross-modal representation learning and improve alignment performance \cite{large,unify,diff}. 

However, directly applying these cross-modal alignment strategies to 3D human motion remains challenging. Human motion involves complex spatial dependencies between body joints and strong temporal correlations across frames. Moreover, natural language descriptions of motion often include multiple actions that occur sequentially or simultaneously across different body parts. Existing alignment frameworks rarely consider these structured dependencies, making it difficult to achieve precise correspondence between language and motion at a fine-grained spatial level.


\begin{figure*}
    \centering
\includegraphics[width=1\linewidth]{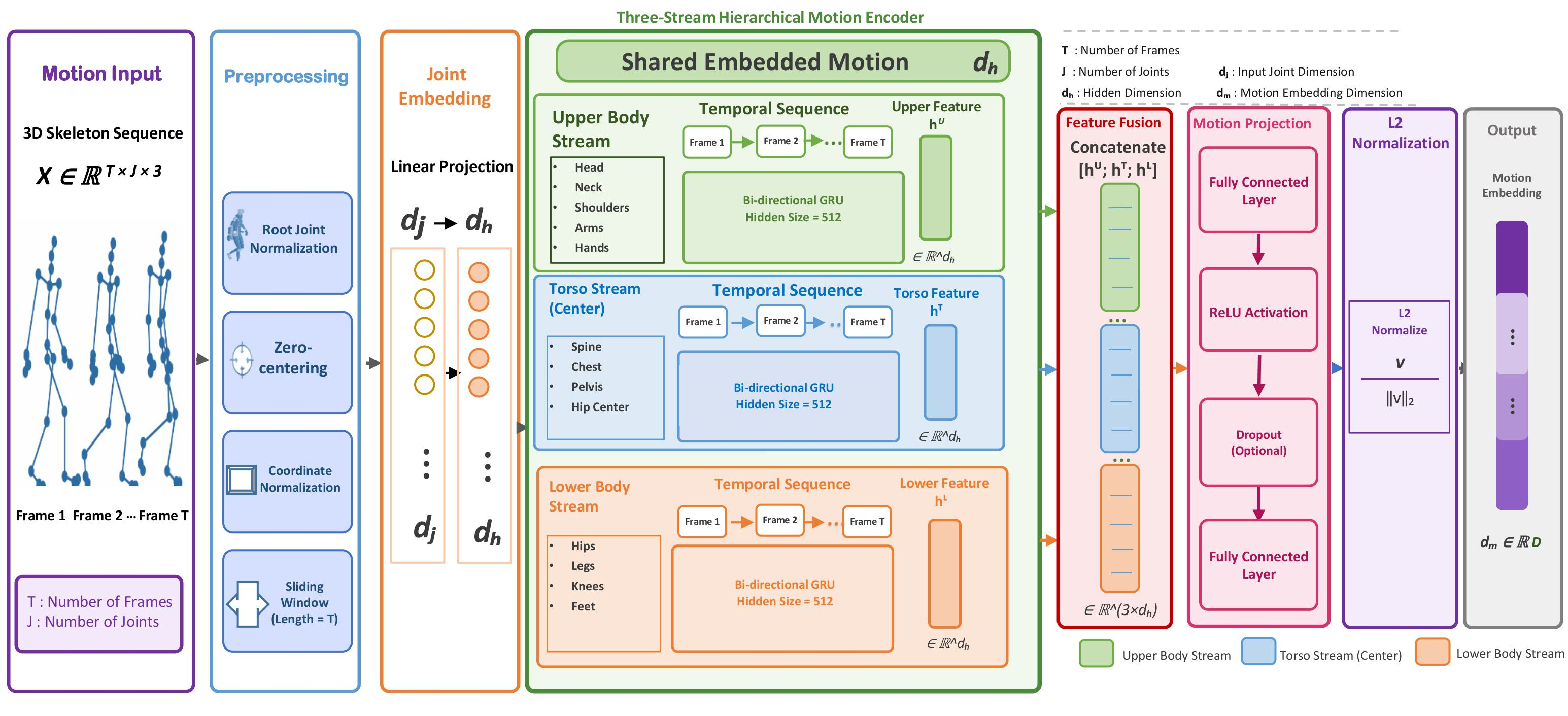}
    \caption{Illustration of the HUMAN-TCI motion pipeline. The input 3D skeletal sequence is first preprocessed and reduced from all joints (21-kitml, 22 H-ml3d) to 5 anatomically meaningful body regions, which are then organized into three body-part streams: upper body, torso, and lower body. The three streams are projected into a shared feature space and passed to the Torso-Center Interaction (TCI) module, detailed in Figure~3, where torso-aware attention models cross-stream dependencies between the upper-body, torso, and lower-body representations. The resulting attended features are then independently encoded using GRU layers to capture temporal motion patterns. Finally, the encoded upper-body, torso, and lower-body features are concatenated and passed through a nonlinear projection layer followed by $L_2$ normalization to obtain the final motion embedding for text-to-motion retrieval.}

    \label{fig:MT}
\end{figure*}

Therefore, a retrieval framework that explicitly models hierarchical body-part interactions and spatial dependencies between torso, upper-body, and lower-body motions is essential for achieving accurate and interpretable text-to-motion retrieval.
\section{Methodology}

\subsection{Problem Formulation}

We study the TMR task, where the goal is to retrieve a semantically consistent human motion sequence $\mathbf{M}$ from a natural language description $\mathbf{W}$. Formally, let $\mathbf{M} = \{ \mathbf{m}_1, \mathbf{m}_2, \dots, \mathbf{m}_T \}$ denote a 3D motion sequence of length $T$, where each frame $\mathbf{m}_t \in \mathbb{R}^{J \times 3}$ encodes the 3D positions of $J$ joints. Similarly, $\mathbf{W} = \{ w_1, w_2, \dots, w_S \}$ represents a textual description of length $S$. The objective is to learn embedding functions $f_m(\cdot)$ and $f_t(\cdot)$ that map motion sequences and textual descriptions into a shared feature space $\mathcal{Z}$, such that semantically related pairs $(\mathbf{W}, \mathbf{M})$ are assigned higher similarity scores than non-matching pairs:
\begin{equation}
\operatorname{sim}\!\left(f_t(\mathbf{W}),f_m(\mathbf{M})\right)
>
\operatorname{sim}\!\left(f_t(\mathbf{W}),f_m(\mathbf{M}')\right).
\end{equation}
Here, $\mathbf{M}$ denotes the ground-truth motion corresponding to the text description $\mathbf{W}$, whereas $\mathbf{M}'$ represents a non-matching (negative) motion sample. The function $\operatorname{sim}(\cdot,\cdot)$ denotes cosine similarity, which measures the similarity between text and motion embeddings in the shared embedding space as shown in Figure~\ref{fig:MT} . The contrastive objective is employed to learn this shared space by increasing the similarity between semantically corresponding text to motion pairs while decreasing the similarity between mismatched pairs, thereby improving text to motion retrieval. Datasets such as HumanML3D and KIT-ML contain multiple textual descriptions for the same motion sequence, resulting in a one-to-many correspondence between language and motion. During training, we treat each text-motion pair as an independent positive instance, while other samples in the batch serve as negatives, following a standard contrastive learning paradigm.

\subsection{Hierarchical Multi-Stream Motion-Aware Network}

Our \textbf{HUMAN} decomposes human motion into three higher-level streams: the upper-body motion stream $\mathbf{M}_\text{upper}$, the torso motion stream $\mathbf{M}_\text{torso}$, and the lower-body motion stream $\mathbf{M}_\text{lower}$. Each stream is constructed from anatomically defined joint groups and processed through torso-centered cross-stream interaction before temporal encoding. Rather than explicitly fusing independently obtained stream-level embeddings, the proposed architecture integrates inter-stream dependencies through torso-guided attention, then combines the temporally encoded streams to form a unified motion representation. The final motion embedding used for text to motion retrieval is expressed as
\begin{equation}
\mathbf{z}_{\mathrm{motion}}
=
f_{\mathrm{motion}}
\left(
\mathbf{M}_{\mathrm{upper}},
\mathbf{M}_{\mathrm{torso}},
\mathbf{M}_{\mathrm{lower}}
\right),
\end{equation}
where $f_{\mathrm{motion}}(\cdot)$ denotes the proposed hierarchical multi-stream motion encoder with torso-centered cross-stream attention, and $\mathbf{M}_{\mathrm{upper}}$, $\mathbf{M}_{\mathrm{torso}}$, and $\mathbf{M}_{\mathrm{lower}}$ represent the upper-body, torso, and lower-body motion streams, respectively.

\subsection{Multi-Stream Motion Encoding}

Each input motion sequence is represented as a sequence of full-body skeletal poses, comprising 21 joints for KIT-ML and 22 joints for HumanML3D. The joints are first organized into five anatomically defined groups: right arm, left arm, right leg, left leg, and mid-body. Each group is independently projected into a part-specific feature space at the frame level. Torso-aware cross-stream interaction is then applied, in which each of the four limb streams (right arm, left arm, right leg, and left leg) attends to the mid-body stream. This interaction allows the network to model spatial relationships between the limbs and the torso before temporal encoding. The torso-informed right and left arm features are then concatenated to form the upper-body stream, while the torso-informed right and left leg features are concatenated to form the lower-body stream. Together with the mid-body stream, this reduces the five anatomical groups to three higher-level streams: upper body, lower body, and torso. Each of these three streams is subsequently processed by a dedicated GRU-based temporal encoder, and the resulting representations are concatenated and projected to obtain the final motion representation.
\begin{figure}[]
    \centering
    \includegraphics[width=\linewidth]{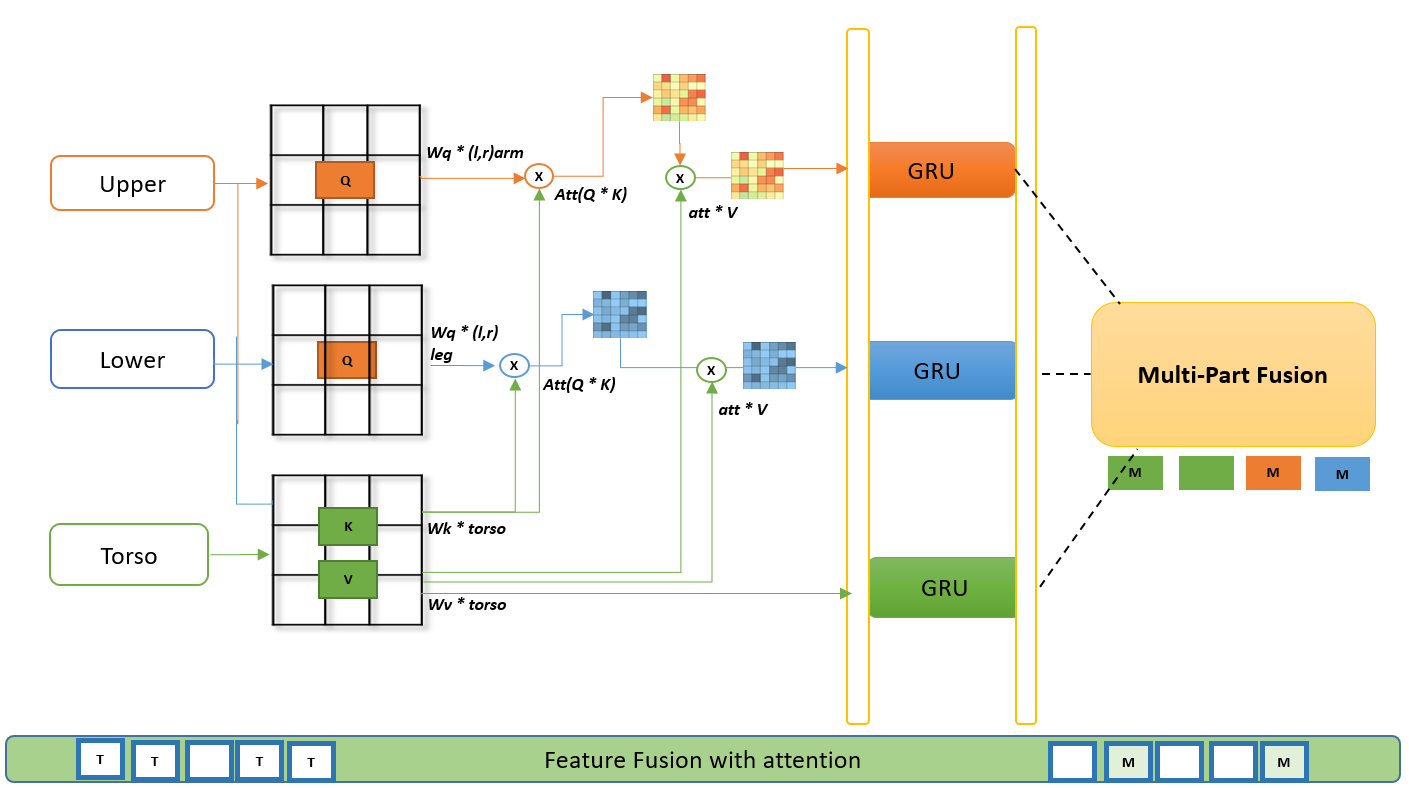}
    \caption{HUMAN-TCI module. Motion is decomposed into upper, lower, and torso streams. Upper/lower streams form queries, while the torso provides keys/values for attention. The attended features are encoded with GRU layers and fused into a unified motion embedding.}
  \label{fig:att}
\end{figure}
\subsubsection{Body-Part Motion Decomposition}
Let a motion sequence $\mathbf{M} \in \mathbb{R}^{T \times J \times 3}$
consist of $T$ frames and $J$ skeletal joints. The joints are first divided into five anatomically defined groups: left arm, right arm, left leg, right leg, and mid-body. These groups are then organized into three motion streams: upper body, torso, and lower body. Specifically, the upper-body stream combines the left- and right-arm joints, the torso stream contains the mid-body joints, and the lower-body stream combines the left- and right-leg joints.
\begin{equation}
\begin{aligned}
\mathbf{J}_{\text{upper}}
&=
\mathbf{J}_{\text{left-arm}}
\cup
\mathbf{J}_{\text{right-arm}},\\
\mathbf{J}_{\text{torso}}
&=
\mathbf{J}_{\text{mid-body}},\\
\mathbf{J}_{\text{lower}}
&=
\mathbf{J}_{\text{left-leg}}
\cup
\mathbf{J}_{\text{right-leg}}.
\end{aligned}
\end{equation}
For HumanML3D, the corresponding joint groups are
$\mathbf{J}_{\text{upper}}\\=\{13,14,16,17,18,19,20,21\}$,
$\mathbf{J}_{\text{torso}}=\{0,3,6,9,12,15\}$, and
$\mathbf{J}_{\text{lower}}=\{1,2,4,5,7,8,10,11\}$.
For KIT-ML, they are
$\mathbf{J}_{\text{upper}}=\{5,6,7,8,9,10\}$,
$\mathbf{J}_{\text{torso}}=\{0,1,2,3,4\}$, and
$\mathbf{J}_{\text{lower}}=\{11,12,13,14,15,16,17,18,19,20\}$.
The indices follow the zero-based joint indexing used in the implementation.

We extract each body-part motion stream individually. Let
$\text{part}\in\{\text{upper},\text{torso},\text{lower}\}$
denote the corresponding body part. The motion representation for each part is obtained by selecting the joints associated with $\mathbf{J}_{\text{part}}$ along the joint dimension:
\begin{equation}
\mathbf{M}_{\text{part}}
=
\mathbf{M}[:,\mathbf{J}_{\text{part}},:].
\end{equation}
This decomposition separates the motion into anatomically meaningful streams while preserving the temporal ordering of the original motion sequence.

\subsubsection{Torso-Guided Attention Mechanism}
\label{att1}

Before temporal encoding, each body-part group is projected into a common feature space. Let
$\mathbf{X}_{\text{upper}}$,
$\mathbf{X}_{\text{lower}}$, and
$\mathbf{X}_{\text{torso}}$
denote the resulting frame-level feature representations. The upper- and lower-body streams generate query representations, while the torso features serve as keys and values. This design enables the limb streams to selectively incorporate contextual information from torso dynamics while preserving their own motion characteristics. The torso-guided attention operations are defined as
\begin{equation}
\begin{aligned}
\mathbf{X}'_{\text{upper}}
&=
\operatorname{Attn}
\left(
\mathbf{Q}_{\text{upper}},
\mathbf{K}_{\text{torso}},
\mathbf{V}_{\text{torso}}
\right),
\label{eq:upper_attention}\\
\mathbf{X}'_{\text{lower}}
&=
\operatorname{Attn}
\left(
\mathbf{Q}_{\text{lower}},
\mathbf{K}_{\text{torso}},
\mathbf{V}_{\text{torso}}
\right).
\end{aligned}
\end{equation}
where the scaled dot-product attention is given by
\begin{equation}
\operatorname{Attn}(Q,K,V)
=
\operatorname{Softmax}
\left(
\frac{QK^{\top}}{\sqrt{d}}
\right)V.
\label{eq:scaled_attention}
\end{equation}
The torso stream does not undergo an additional attention update and is therefore retained as
\begin{equation}
\mathbf{X}'_{\text{torso}}
=
\mathbf{X}_{\text{torso}}.
\label{eq:torso_identity}
\end{equation}
Thus, the resulting three torso-aware representations
$\mathbf{X}'_{\text{upper}}$,
$\mathbf{X}'_{\text{torso}}$, and
$\mathbf{X}'_{\text{lower}}$
are passed to the temporal encoding stage. This asymmetric interaction enables the torso to provide contextual guidance to the upper- and lower-body streams without introducing an additional self-attention update for the torso stream as shown in Figure~\ref{fig:att}.

\subsubsection{Temporal Modeling and Fusion}
The torso-aware representations are subsequently processed by stream-specific Gated Recurrent Unit (GRU) encoders to model temporal dependencies across the $T$ frames:
\vspace{-2mm}
\begin{equation}
\mathbf{H}_{\text{part}}
=
\operatorname{GRU}_{\text{part}}
\left(
\mathbf{X}'_{\text{part}}
\right)
\in
\mathbb{R}^{T\times d},
\end{equation}
where $\text{part}\in\{\text{upper},\text{torso},\text{lower}\}$. $T$ denotes the number of frames (i.e., the temporal sequence length) and $d$ denotes the GRU feature dimension. The stream-specific GRU encoders capture the temporal evolution of the torso-informed upper-body and lower-body representations while preserving the original torso dynamics.

The resulting stream-level representations are then integrated using a multi-part fusion operation:
\begin{equation}
\mathbf{Z}_{\text{motion}}
=
\operatorname{Fuse}
\left(
\mathbf{H}_{\text{upper}},
\mathbf{H}_{\text{torso}},
\mathbf{H}_{\text{lower}}
\right),
\end{equation}
where $\operatorname{Fuse}(\cdot)$ denotes concatenation followed by a learnable projection layer. The resulting representation $\mathbf{Z}_{\text{motion}}$ is subsequently mapped to the shared motion--text embedding space and used for text--motion retrieval. Through this hierarchical design, HUMAN-TCI captures temporal dynamics within individual body parts while explicitly modeling torso-guided interactions between the upper and lower body.

\begin{figure*}
    \centering
    \includegraphics[width=1\linewidth,height=6cm]{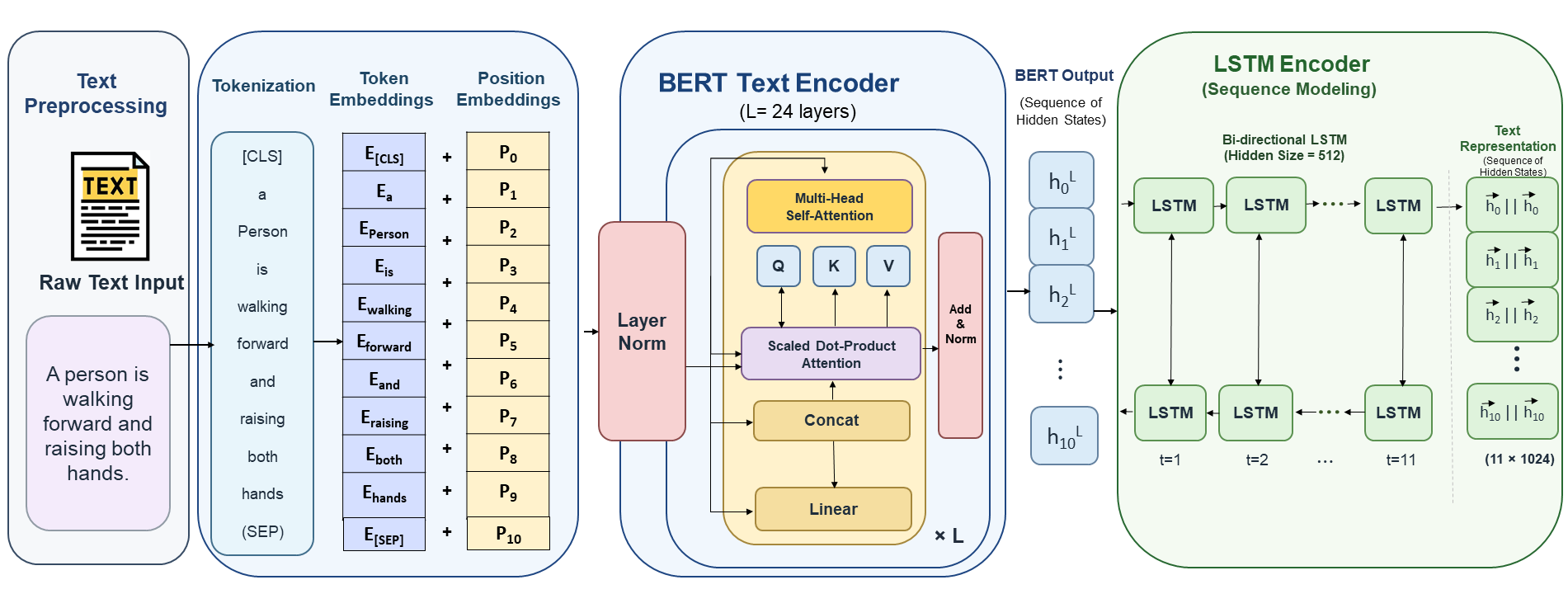}
    \caption{ Detail proposed text encoding pipeline in HUMAN-TCI. A natural language description is first tokenized and encoded using BERT to obtain contextualized token-level hidden representations. The resulting sequence of BERT hidden states is then passed to a bidirectional LSTM to model sequential dependencies and capture long-range contextual information. The resulting text representation is subsequently projected into the shared motion--text embedding space, where it is aligned with the corresponding motion representation for fine-grained text-to-motion retrieval.}
    \label{fig:4}
\end{figure*}

\subsection{Text Encoding Module}

To enable motion retrieval from natural language descriptions, HUMAN-TCI learns a semantic text representation that captures both linguistic meaning and temporal action structure. As illustrated in Figure~4, the text encoding pipeline consists of two stages: 1) contextual word representation using a pretrained language model and 2) sequential modeling using a recurrent encoder.

\subsubsection{Contextual Word Embeddings}

Given a text description $\mathbf{W} = \{w_1, w_2, \ldots, w_S\}$ of length $S$, we first obtain contextualized token representations using a pretrained BERT-Large-Cased model as the backbone for textual feature extraction. Instead of relying solely on the final layer, we concatenate hidden states from multiple upper layers to capture rich semantic and syntactic information.

Formally, let $\mathbf{E}^{(l)} \in \mathbb{R}^{S \times d}$ denote the hidden states from layer $l$. We select the four layers from 12 to 15 and construct the word embedding matrix as:
\begin{equation}
\mathbf{E}_\text{BERT} =
\mathrm{Concat}\big(\mathbf{E}^{(L-i)}\big)_{i=12}^{15}
\in \mathbb{R}^{S \times d'},
\end{equation}
where $L$ is the total number of layers and $d' = 4d$. This multi-layer aggregation captures both high-level semantics and intermediate linguistic features, which helps describe complex human actions.

\subsubsection{Sequential Sentence Encoding}

Natural language descriptions of motion often contain temporal cues (e.g., “walking while waving”), requiring sequential modeling beyond static embeddings. Therefore, the contextual token embeddings are processed by a multi-layer Long Short-Term Memory (LSTM) network:
\begin{equation}
\mathbf{h}_t = \text{LSTM}(\mathbf{E}_\text{BERT}(t), \mathbf{h}_{t-1}).
\end{equation}
After processing the full sequence, the final hidden state of the top LSTM layer is taken as the sentence representation:
\begin{equation}
\mathbf{z}_\text{text}^{raw} = \mathbf{h}_S.
\end{equation}
To ensure compatibility with the hierarchical motion representation, the hidden dimension of the LSTM is set to three times the base embedding size, enabling the model to implicitly encode information relevant to upper-body, lower-body, and torso motions.

\subsubsection{Projection to the Joint Embedding Space}

The raw sentence embedding is further projected into the shared motion and text embedding space using a linear transformation:
\begin{equation}
\mathbf{z}_\text{text} =
\mathbf{F}_p \mathbf{z}_\text{text}^{raw} + \mathbf{b}_p,
\end{equation}
where $\mathbf{F}_p$ and $\mathbf{b}_p$ are learnable parameters.

\subsubsection{CLIP Text Encoder.}
We also experiment with a pretrained Contrastive Language Image Pre-training (CLIP) text encoder to evaluate the impact of large-scale vision--language pretraining on motion retrieval. Specifically, we use the pretrained \texttt{ViT-B/32} CLIP model, with the text encoder initialized from its publicly released pretrained weights. The CLIP text encoder is kept frozen during training, such that only the subsequent projection layer and the motion-side components are optimized. Input sentences are tokenized using the CLIP tokenizer with the model's maximum supported context length of 77 tokens; sequences exceeding this limit are truncated, while shorter sequences are padded according to the CLIP implementation. The resulting 512-dimensional CLIP text representation is then passed through the same text projection head used in the baseline configuration and mapped into the shared 256-dimensional motion--text embedding space. This configuration keeps the downstream architecture and embedding dimensionality unchanged, allowing us to evaluate the effect of the CLIP-based textual representation independently. The resulting variant therefore assesses HUMAN-TCI's robustness to alternative pretrained textual representations while maintaining a consistent retrieval framework.

\subsection{Model Training and Loss Function}

Hierarchical motion embeddings $Z_\text{motion}$ from the multi-stream motion encoder and projected text embeddings $z_\text{text}$ are mapped into a shared embedding space $\mathcal{Z}$:
\begin{equation}
z_\text{motion}, z_\text{text} \in \mathcal{Z} \subset \mathbb{R}^{d}.
\end{equation}

The similarity between a text query and a candidate motion sequence is computed using cosine similarity:
\begin{equation}
S(z_\text{text}, Z_\text{motion}) = \frac{z_\text{text} \cdot Z_\text{motion}}{\|z_\text{text}\| \, \|Z_\text{motion}\|}.
\end{equation}

We explore two widely-used metric learning objectives: symmetric triplet loss and InfoNCE loss. Both aim to bring matching text--motion pairs closer while separating non-matching pairs in the shared embedding space.

\paragraph{Symmetric Triplet Loss}
Given a batch of size $B$ with paired embeddings $\{(z_{\text{text},i}, Z_{\text{motion},i})\}$, the symmetric triplet loss is defined as:
\begin{equation}
\begin{aligned}
\mathcal{L}_\text{Triplet} = \frac{1}{B} \sum_{i=1}^{B} 
\Big[
&\max_{j \neq i} \big( \alpha + S(z_{\text{text},i}, Z_{\text{motion},j}) \\
&\quad - S(z_{\text{text},i}, Z_{\text{motion},i}) \big)_+ \\
&+ \max_{j \neq i} \big( \alpha + S(z_{\text{text},j}, Z_{\text{motion},i}) \\
&\quad - S(z_{\text{text},i}, Z_{\text{motion},i}) \big)
\Big],
\end{aligned}
\end{equation}
where $(x)_+ = \max(0, x)$ and $\alpha$ is a margin hyperparameter. The index $j$ corresponds to the hardest negative sample in the batch.
\begin{table*}[!t]
\centering
\caption{Comparison with state-of-the-art methods on KIT-ML \cite{Kit} and HumanML3D \cite{Gen}.}
\label{tab:combined_t2m}

\resizebox{\textwidth}{!}{%
\begin{tabular}{lcccccccccc}
\toprule
\textbf{Method} & \textbf{Pub. Year} &
\multicolumn{4}{c}{\textbf{KIT-ML}} &
\multicolumn{4}{c}{\textbf{HumanML3D}} \\
\cmidrule(lr){3-6}
\cmidrule(lr){7-10}
&
&
\textbf{R@1 $\uparrow$} &
\textbf{R@5 $\uparrow$} &
\textbf{R@10 $\uparrow$} &
\textbf{MedR $\downarrow$} &
\textbf{R@1 $\uparrow$} &
\textbf{R@5 $\uparrow$} &
\textbf{R@10 $\uparrow$} &
\textbf{MedR $\downarrow$} \\
\midrule

T2M \cite{Gen} & CVPR'22 &
3.37 & 16.87 & 27.71 & 28 &
1.80 & 7.12 & 12.47 & 81 \\

MotionCLIP \cite{m_clip} & ECCV'22 &
4.87 & 20.09 & 31.57 & 26 &
2.33 & 12.77 & 18.14 & 103 \\

TEMOS \cite{temos} & ECCV'22 &
7.11 & 24.10 & 35.66 & 24 &
2.12 & 8.26 & 13.52 & 173 \\

MoT \cite{3} & SIGIR'23 &
6.23 & 23.92 & 37.15 & 20 &
2.61 & 10.66 & 17.79 & 60 \\

TMR \cite{petrovich2023tmr} & ICCV'23 &
7.23 & 28.31 & 40.12 & 17 &
5.68 & 20.34 & 30.94 & 28 \\

HSA \cite{h1} & SIGIR'24 &
9.29 & 29.01 & 40.97 & 16 &
7.14 & 24.02 & 34.67 & 24 \\

MGSI \cite{h4} & MM'24 &
8.91 & 29.64 & 40.84 & 16 &
6.61 & 23.91 & 34.74 & 24 \\

Messi-B \cite{3} & SIGIR'23 &
3.20 & 15.70 & 25.30 & 34 &
2.40 & 10.50 & 17.70 & 68 \\

DTL \cite{yan2023cross} & MM'23 &
6.77 & 23.18 & 37.24 & 18 &
2.30 & 10.06 & 16.40 & 76 \\

RetNet \cite{transformer} & PatR'26 &
9.59 & 30.56 & 43.07 & 15 &
7.61 & 25.65 & 35.04 & 24 \\

\textbf{HUMAN-TCI (Ours)} & -- &
\textbf{9.96} & \textbf{32.31} & \textbf{47.07} & \textbf{13} &
\textbf{8.21} & \textbf{27.17} & \textbf{38.87} & \textbf{16} \\

\bottomrule
\end{tabular}%
}

\begin{tablenotes}
\footnotesize
\item Arrows indicate whether higher ($\uparrow$) or lower ($\downarrow$) values are better. Bold indicates the best performance in each column.
\end{tablenotes}

\end{table*}
\paragraph{InfoNCE Loss}
We adopt the InfoNCE loss as our primary training objective due to its superior performance. It is formulated as a symmetric cross-entropy loss:
\begin{equation}
\begin{aligned}
\mathcal{L}_\text{InfoNCE} = - \frac{1}{B} \sum_{i=1}^{B}
\Bigg[
&\log \frac{\exp(S(z_{\text{text},i}, Z_{\text{motion},i}) / \tau)}
{\sum_{j=1}^{B} \exp(S(z_{\text{text},i}, Z_{\text{motion},j}) / \tau)} \\
&+ \log \frac{\exp(S(z_{\text{text},i}, Z_{\text{motion},i}) / \tau)}
{\sum_{j=1}^{B} \exp(S(z_{\text{text},j}, Z_{\text{motion},i}) / \tau)}
\Bigg],
\end{aligned}
\end{equation}
where $\tau$ is a temperature parameter. During inference, we encode a text query and compare it against all motion embeddings using cosine similarity. We retrieve the top-$k$ motions based on similarity scores.

\section{Experiments}

\subsection{Datasets}

We evaluate HUMAN-TCI on two widely used human motion retrieval datasets: HumanML3D~\cite{Gen} and KIT Motion Language (KIT-MoCap)~\cite{Kit}, both providing one or more textual descriptions per motion sequence.

Data Representation: In both datasets, each joint is represented with $D=9$ features: six for continuous rotation and three for rotation-invariant forward-kinematics joint positions. Preprocessing follows the HumanML3D pipeline~\cite {Gen}.

KIT Motion-Language Dataset: KIT-MoCap contains 3,911 full-body motions in Master Motor Map (MMM) format~\cite{Kit} with 6,278 English textual annotations. Each motion may have multiple descriptions (e.g., ``A human walks two steps forwards, pivots 180 degrees, and walks two steps back''). For evaluation, 938 textual queries are used to search among 734 motion sequences after removing redundant queries.

HumanML3D Dataset: HumanML3D \cite{Gen} builds on - AMASS and HumanAct12, with additional textual annotations, totaling 14,616 motion sequences and 44,970 descriptions. For retrieval, 8,401 textual queries search among 4,198 motion sequences. Since descriptions often correspond to motion subsequences, this dataset enables finer-grained retrieval of specific motion segments.

These datasets provide paired text and human motion sequences, covering a diverse set of actions and natural language descriptions. In particular, HumanML3D includes many compositional and multi-action descriptions, making it well-suited for evaluating fine-grained text–motion alignment. Similarly, KIT includes a wide range of motion categories with varying complexity, enabling comprehensive evaluation of retrieval performance.

An analysis of these datasets reveals that over 90\% of motion sequences involve multiple action events, and their corresponding textual descriptions often include sequential or overlapping actions \cite{1}. Many descriptions explicitly reference torso-related movements such as bending, twisting, or turning, highlighting the importance of fine-grained spatial modeling in text-guided motion retrieval.

\subsection{Experimental setup}

The proposed HUMAN-TCI model aligns textual descriptions and motion sequences in a shared embedding space. The motion encoder is a hierarchical, multi-stream GRU network (Upper-Lower-Torso GRU), while the text encoder uses either BERT-Large-Cased followed by a multi-layer LSTM or CLIP’s pretrained text encoder. We experimented with both InfoNCE and symmetric triplet losses; results reported here use InfoNCE due to superior performance.

\subsubsection{Implementation Details}

Training is performed using the Adam optimizer with a learning rate of $3 \times 10^{-5}$ and a batch size of 96. For the KIT-ML dataset, the
model is trained for 120 epochs using a cosine annealing learning-rate
scheduler with $T_{\max}=120$. For HumanML3D, the model is trained for
30 epochs using a MultiStepLR scheduler with a milestone at epoch 20 and
a decay factor of $\gamma=0.1$. The random seed is fixed to 42 to ensure
reproducibility. A dropout rate of 0.2 is applied to the final projection
layers to improve generalization, and the shared embedding space is set
to a dimensionality of 256.

The model is trained using a contrastive learning objective based on the InfoNCE loss, with a temperature of 0.07. The loss configuration uses a maximum logit scale of 100 and a margin of 0.01, which is applied to the positive matching pairs. We further apply label smoothing with a factor of 0.1 to both motion-to-text and text-to-motion contrastive objectives. Both motion and text embeddings are $\ell_2$-normalized prior to similarity computation, enabling cosine-similarity-based retrieval. Bidirectional matching is enforced by computing the contrastive objective in both motion-to-text and text-to-motion directions. We also incorporate a hard-negative penalty (HNP) weighted by 0.05, selecting the highest-scoring non-matching sample in the mini-batch as the hardest negative. A margin-based penalty is applied when its similarity approaches the positive pair, encouraging better separation between positive and challenging negative samples.

For data processing, motion sequences are represented using a continuous
6D rotation representation combined with RIFKE features. For KIT-ML,
fixed-length clips of 50 frames are used, whereas HumanML3D uses the
original variable sequence lengths. The same data-loading configuration
is adopted across the training, validation, and testing splits, with zero
worker threads to ensure deterministic and reproducible data loading.
The maximum-violation setting is 5 for KIT-ML and 4 for HumanML3D.

Validation is performed periodically during training to monitor retrieval
performance. We conduct validation every 100 training iterations.
Performance is evaluated using Recall@K (R@1, R@5, and R@10), median rank
(MedR), mean rank (MeanR), and semantic relevance measured using SpaCy
similarity. The Hydra framework is utilized to provide flexible and
modular configuration of model architectures, optimization strategies,
and dataset parameters, enabling systematic experimentation across
different model and dataset configurations.

\subsubsection{Retrieval and Evaluation}
During inference, a textual query is encoded and compared with all precomputed motion embeddings using cosine similarity, and the top-$k$ most similar motions are retrieved. We evaluate performance using standard retrieval metrics, including Recall@$K$ (R@$K$), Median Rank (MedR), and Mean Rank (MeanR). To further assess semantic alignment, we employ Normalized Discounted Cumulative Gain (NDCG) SPICE, and spaCy-based similarity. Additionally, qualitative visualizations are used to analyze the correspondence between textual descriptions and body-part-specific motion dynamics.

\subsection{Performance Comparison}

Table~1 presents a comprehensive comparison with state-of-the-art methods on the KIT-ML and HumanML3D datasets. Overall, our method achieves consistently strong performance across both datasets, attaining the best results across all reported retrieval metrics while maintaining a lightweight and efficient architecture (Table~4).
On the KIT-ML dataset, our method achieves the best performance across all metrics, obtaining an R@1 of 9.96, R@5 of 32.31, an R@10 of 47.08, and a MedR of 13. In particular, the substantial gains in R@5 and R@10 demonstrate improved retrieval consistency at higher recall levels, indicating stronger ranking quality and robustness in capturing diverse motion variations. In comparison, complex transformer-based methods such as TMR~\cite{petrovich2023tmr}, RetNet~\cite{transformer}, and MoT~\cite{3} our method achieve better performance across the reported recall and median-rank metrics.

\begin{table}[b]
\centering
\caption{Ablation study on KIT and HumanML3D datasets.}
\label{tab:ablation_main}
\scriptsize  
\setlength{\tabcolsep}{4pt} 
\resizebox{\columnwidth}{!}{%
\begin{tabular}{l c c c c c}
\toprule
\textbf{Motion Model} & \textbf{R@1 $\uparrow$} & \textbf{R@5 $\uparrow$} & \textbf{R@10 $\uparrow$} & \textbf{MedR $\downarrow$} & \textbf{MeanR $\downarrow$} \\
\midrule

\multicolumn{6}{c}{\textbf{KIT }} \\
\midrule
Hier-2TGRU & 3.71 & 15.22 & 28.57 & 31.00 & 72.34    \\
Hier-3TGRU & 5.44 & 21.83 & 36.20 & 21.00 & 61.54 \\
Hier-3TGRU-Att  & 7.86 & 29.65 & 44.48 & 14.00 & 26.97 \\
Hier-3TGRU-Att-HNP & \textbf{9.96} & \textbf{32.31} & \textbf{47.07} & \textbf{13.00} & \textbf{18.51} \\
\midrule
\multicolumn{6}{c}{\textbf{HumanML3D }} \\
\midrule
Hier-2TGRU & 2.42 & 10.56 & 21.00 & 61.00 & 285.7 \\
Hier-3TGRU & 4.67 & 16.43 & 28.65 & 38.00  & 138.0 \\
Hier-3TGRU-Att & 6.29 & 26.37 & 33.56 & 18.00 & 46.27 \\
Hier-3TGRU-Att-HNP & \textbf{8.21} & \textbf{27.17} & \textbf{38.87} & \textbf{16.00} & \textbf{35.72} \\
\bottomrule
\end{tabular}%
}
\end{table}
On the HumanML3D dataset, our method similarly achieves the best performance across all reported metrics, with R@1 of 8.21, R@5 of 27.17, R@10 of 38.87, and MedR of 16. These results demonstrate strong generalization to datasets with more complex and diverse motion descriptions. The improvement in higher-recall metrics further indicates that the proposed model can effectively retrieve relevant motions across a broader range of textual descriptions.

Compared to earlier methods such as T2M~\cite{Gen}, MotionCLIP~\cite{m_clip}, and TEMOS~\cite{temos}, our method shows clear improvements across all reported retrieval metrics, particularly in R@5 and R@10, highlighting its ability to retrieve more relevant candidates within the top-ranked results. Furthermore, compared to sequential baselines such as Messi-B~\cite{3}, as well as other recent state-of-the-art approaches~\cite{yan2023cross,Gen}, our method achieves more favorable performance gains across both datasets.
These improvements stem from our hierarchical multi-stream design, which explicitly models body-part-specific dynamics while enabling cross-part interactions. In particular, a dedicated torso stream lets the model capture global motion coordination, which prior work often overlooks. This improves alignment between textual descriptions and motion sequences, especially for compositional actions involving multiple body regions.

In contrast, methods that rely on simplified body representations or treat the human body as a single unit struggle to capture fine-grained dependencies, resulting in lower recall performance. Additionally, while transformer-based approaches show strong comparative results, they often incur high computational cost. Our method, based on GRU-based motion encoding and a lightweight text encoding pipeline (BERT-LSTM and CLIP), achieves competitive or superior performance with significantly lower complexity (Table~\ref{tab:complexity_compact}).
Overall, the results demonstrate that explicitly modeling hierarchical body structure and torso-centric interactions within a multi-stream framework, while employing a lightweight architecture such as GRU instead of more complex transformer-based models, achieves a strong balance between retrieval accuracy, robustness, and computational efficiency.


\subsection{Ablation Study and Analysis}

To validate the effectiveness of the proposed HUMAN-TCI framework, we conduct a comprehensive ablation study analyzing architectural design, body-part decomposition, cross-part interactions, and language representations. We evaluate all experiments on the KIT and HumanML3D datasets using standard text-to-motion retrieval metrics.

\subsubsection{Impact of Hierarchical Body-Part Modeling}

We first evaluate the contribution of hierarchical motion decomposition. The simple GRU models the human body using two streams, whereas our proposed Hier-3TGRU adds a torso stream to explicitly capture interactions between upper- and lower-body movements. 

As shown in Table~\ref{tab:ablation_main}, incorporating torso modeling consistently improves retrieval performance across both datasets. This validates our hypothesis that the torso acts as a structural and semantic bridge, enabling better modeling of coordinated and compositional human motions.

\subsubsection{Effect of Cross Body-Part Attention}

To further capture interdependencies between body parts, we introduce a tailored cross-attention mechanism at the body-part level. Unlike prior work that relies on simple concatenation~\cite{3}, our approach explicitly models the torso's influence on both upper- and lower-body streams.

Although the attention variant shows competitive performance (Table~\ref{tab:ablation_main}), the results indicate that structured hierarchical modeling already captures strong dependencies, while cross-attention further refines inter-part relationships. Notably, torso interactions play a central role in improving compositional motion understanding.

\subsubsection{Text Encoder Analysis: BERT vs CLIP}

The comparison between BERT and CLIP is presented in Table~\ref{tab:ablation_bert_clip_compact}. While BERT-based models perform reasonably well, replacing them with CLIP yields substantial improvements across all metrics. This demonstrates that vision-language pretraining enables stronger semantic alignment between textual descriptions and motion representations.

\begin{table}[b]
\centering
\caption{Ablation study on KIT and HumanML3D datasets: BERT-LSTM vs CLIP text models. Best results in bold.}
\label{tab:ablation_bert_clip_compact}
\scriptsize
\setlength{\tabcolsep}{4pt} 
\resizebox{\columnwidth}{!}{%
\begin{tabular}{l c c c c c c c}
\toprule
\textbf{Text Model} 
& \multicolumn{3}{c}{\textbf{KIT}} & \multicolumn{3}{c}{\textbf{HumanML3D}} \\
\cmidrule(lr){2-4} \cmidrule(lr){5-7}
 & \textbf{R@1 $\uparrow$} & \textbf{R@5 $\uparrow$} & \textbf{R@10 $\uparrow$} 
 & \textbf{R@1 $\uparrow$} & \textbf{R@5 $\uparrow$} & \textbf{R@10 $\uparrow$} \\
\midrule
BERT-Large & 7.82 & 29.57 & 44.27 & 6.84 & 25.1 & 36.54\\
CLIP      & \textbf{9.96} & \textbf{32.31} & \textbf{47.08} & \textbf{8.21} & \textbf{27.17} & \textbf{38.87} \\
\bottomrule
\end{tabular}%
}
\end{table}

\subsubsection{Computational Efficiency Analysis}

\begin{table}[b]
\centering
\caption{Computational efficiency comparison of text-to-motion retrieval models on KIT-ML and HumanML3D. Full inference time is reported in minutes.}
\label{tab:complexity_compact}
\scriptsize
\setlength{\tabcolsep}{6pt} 
\begin{tabular}{l c c c c c c}
\toprule
\textbf{Model} & \begin{tabular}{@{}c@{}}\textbf{Body} \\ \textbf{Parts ($J$)}\end{tabular} & \begin{tabular}{@{}c@{}}\textbf{Batch} \\ \textbf{Size}\end{tabular} & \multicolumn{2}{c}{\textbf{Time (m)}} \\
\cmidrule(lr){4-5}
 &  &  & \textbf{KIT-ML} & \textbf{HumanML3D} \\
\midrule
BERT-LSTM + GRU       & 5  & 32 & 4.47 & 4.82  \\
CLIP + GRU            & 5  & 32 & \textbf{2.36} & \textbf{2.51}  \\
Full Joint Model      & 22 & 32 & 6.12 & 6.45  \\
\bottomrule
\end{tabular}
\end{table}
\begin{figure*}[t]
   \includegraphics[width=1\linewidth]{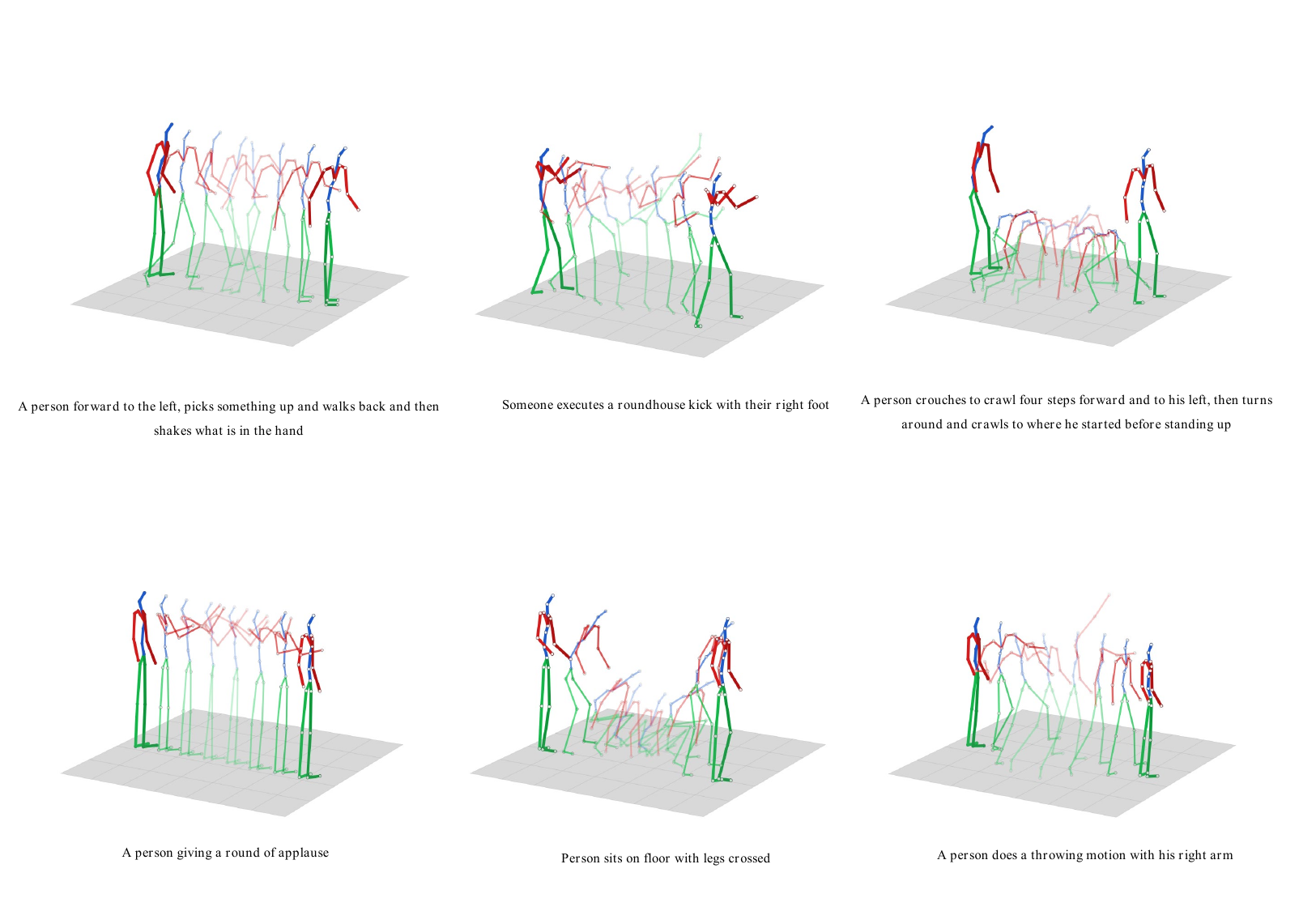}
    \caption{3D skeleton visualization depicting the spatiotemporal evolution of human motion across sequential frames. The visualization represents the temporal progression of articulated body joints, highlighting variations in body posture, joint movements, and structural coordination during motion execution. Such skeletal representations provide a compact view of human dynamics for learning discriminative motion features.}
    \label{fig:Skel5}
\end{figure*}
To improve computational efficiency, we aggregate skeletal joints into five body-part representations ($J=5$), significantly reducing inference time while preserving essential motion dynamics. As shown in Table~\ref{tab:complexity_compact}, this design achieves nearly 2x faster inference than processing all joints ($J\sim22$). At the text encoding level, BERT-LSTM models sequential dependencies and captures fine-grained action descriptions, but its sequential nature introduces additional computational overhead. In contrast, CLIP leverages a parallelizable transformer architecture, providing faster embedding computation and more effective cross-modal alignment. Consequently, CLIP-based models offer a favorable trade-off between speed and retrieval accuracy, making them suitable for real-time or large-scale text-to-motion retrieval applications.

\subsubsection{Semantic Evaluation}

\begin{figure*}[t]
  
    \includegraphics[width=1\linewidth]{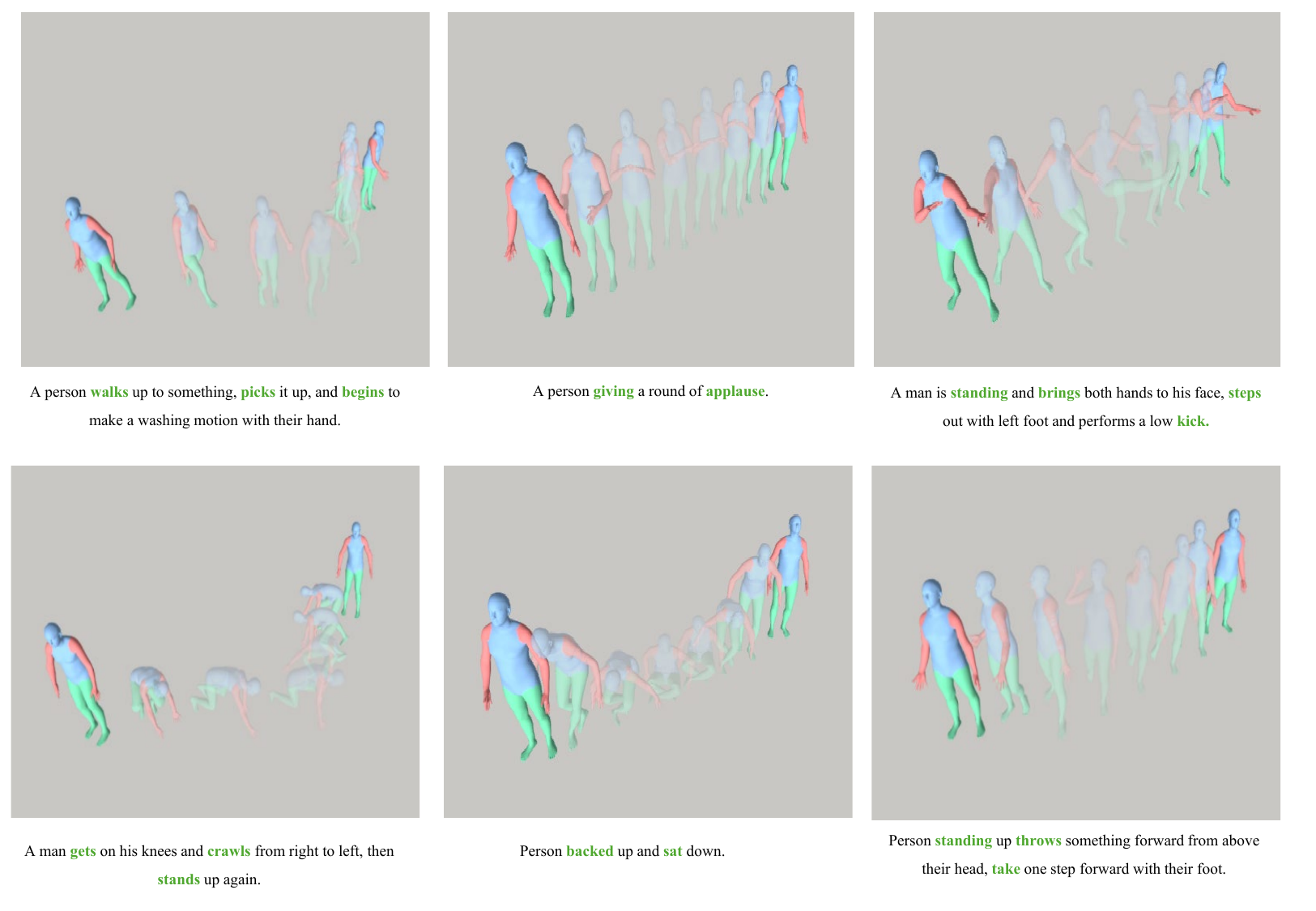}
   \caption{Visualization of human motion using the SMPL body model, showing the reconstructed 3D human mesh over sequential frames. The SMPL-based representation preserves detailed body geometry and pose variations while capturing the spatiotemporal dynamics of articulated human movements, enabling an intuitive interpretation of learned motion patterns.}
    \label{fig:placeholder}
\end{figure*}
Beyond retrieval accuracy, we evaluate semantic alignment using NDCG SPICE, and spaCy similarity. These metrics provide complementary insights into ranking quality, semantic structure, and linguistic similarity. As shown in Table~\ref{tab:semantic_clean}, our model achieves improved semantic consistency, particularly for compositional motion descriptions.

Overall, the ablation study demonstrates that: (i) torso modeling plays a critical role in capturing coordinated motion, (ii) hierarchical decomposition enables fine-grained representation learning, (iii) cross-attention enhances inter-part relationships, and (iv) CLIP, as a vision–language model, significantly improves both performance and efficiency, while BERT-Large effectively captures semantic information in sequential and compositional long descriptions. These findings validate the effectiveness of HUMAN-TCI as a robust and scalable framework for text-to-motion retrieval.
\subsection{Visualization Results}
\begin{figure*}[t]
  
    \includegraphics[width=1\linewidth]{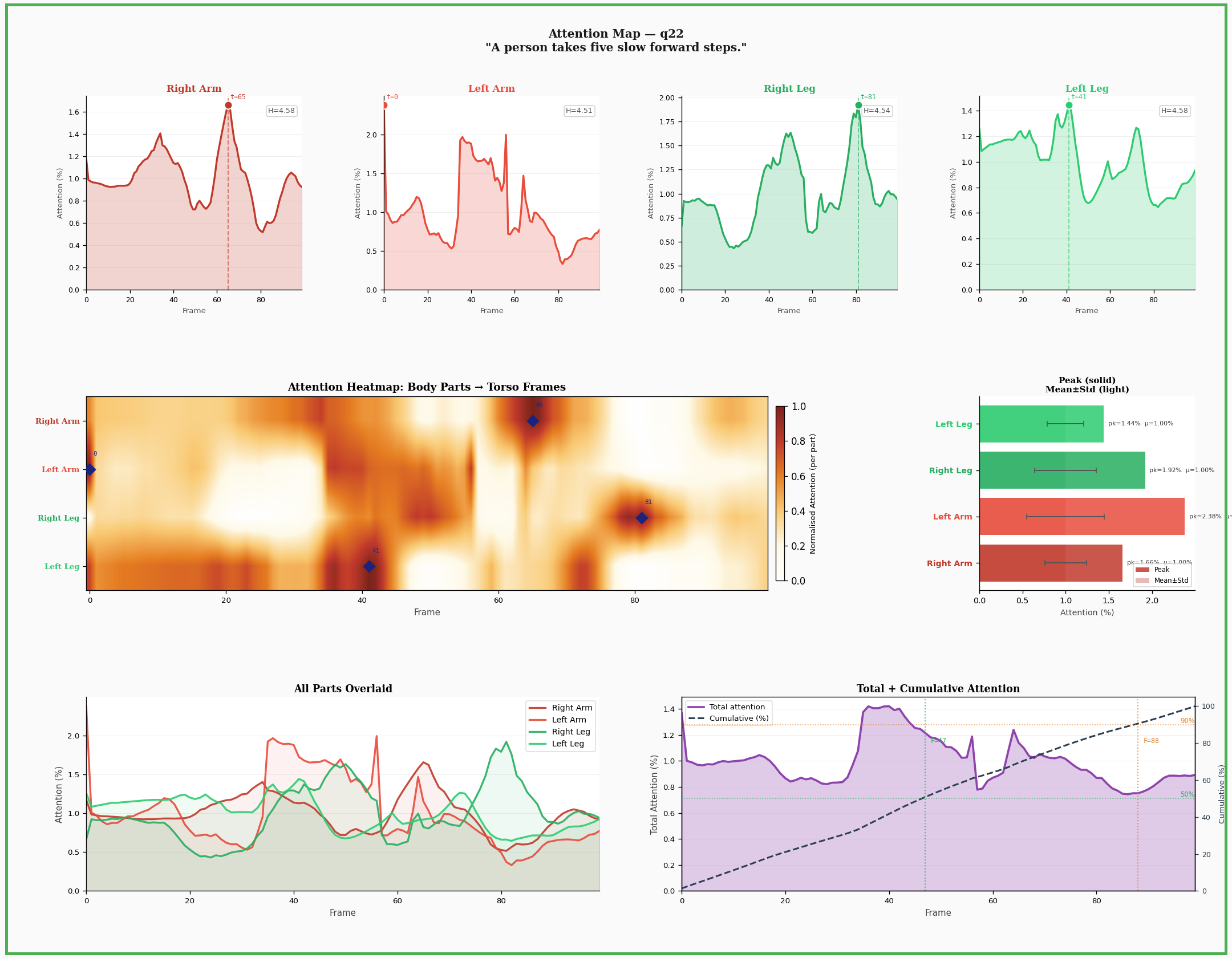}
    \caption{Detailed torso-centered attention analysis for the query
    \textit{``A person takes five slow forward steps.''}
    \textbf{Top row:} per-limb attention distribution over torso frames
    for Right Arm, Left Arm, Right Leg, and Left Leg, with peak attention
    frame ($t$) and Shannon entropy ($H$) annotated for each.
    \textbf{Middle-left:} attention heatmap across all four limbs (rows)
    and torso frames (columns), row-normalized for visual contrast, with
    diamond markers indicating each limb's peak-attention frame.
    \textbf{Middle-right:} peak (solid) versus mean$\pm$std (light)
    attention magnitude per limb.
    \textbf{Bottom-left:} all four limb attention curves overlaid for
    direct comparison.
    \textbf{Bottom-right:} total attention (summed across limbs) and its
    cumulative distribution over the sequence, with $F_{50}$ and $F_{90}$
    marking the frames by which 50\% and 90\% of cumulative attention
    mass is reached, respectively.}
    \label{fig:TCI}
\end{figure*}
We present qualitative results to evaluate the alignment between textual descriptions and retrieved motion sequences. We select a diverse set of queries, including simple, sequential, and compositional actions, to assess the model’s ability to capture fine-grained motion dynamics and coordinated human behavior.
\begin{table}[b]
\centering
\caption{Semantic evaluation of text-to-motion retrieval models on KIT-ML and HumanML3D. Higher values indicate better semantic alignment.}
\label{tab:semantic_clean}
\begin{tabular}{l c c c}
\toprule
\textbf{Model} & \textbf{Dataset} & \multicolumn{2}{c}{\textbf{nDCG $\uparrow$}} \\
\cmidrule(lr){3-4}
 &  & \textbf{SPICE} & \textbf{spaCy} \\
\midrule
Upper-Lower GRU       & KIT-ML       & 0.271 & 0.706 \\
Hier-3TGRU      & KIT-ML       & 0.263 & 0.697 \\
Hier-3TGRU + CLIP     & KIT-ML       & \textbf{0.339} & \textbf{0.770} \\
\midrule
Upper-Lower GRU       & HumanML3D    & 0.318 & 0.723 \\
Hier-3TGRU      & HumanML3D    & 0.316 & 0.729 \\
Hier-3TGRU + CLIP     & HumanML3D    & \textbf{0.355} & \textbf{0.824} \\
\bottomrule
\end{tabular}
\end{table}

\begin{figure*}[t]
  
    \includegraphics[width=1\linewidth]{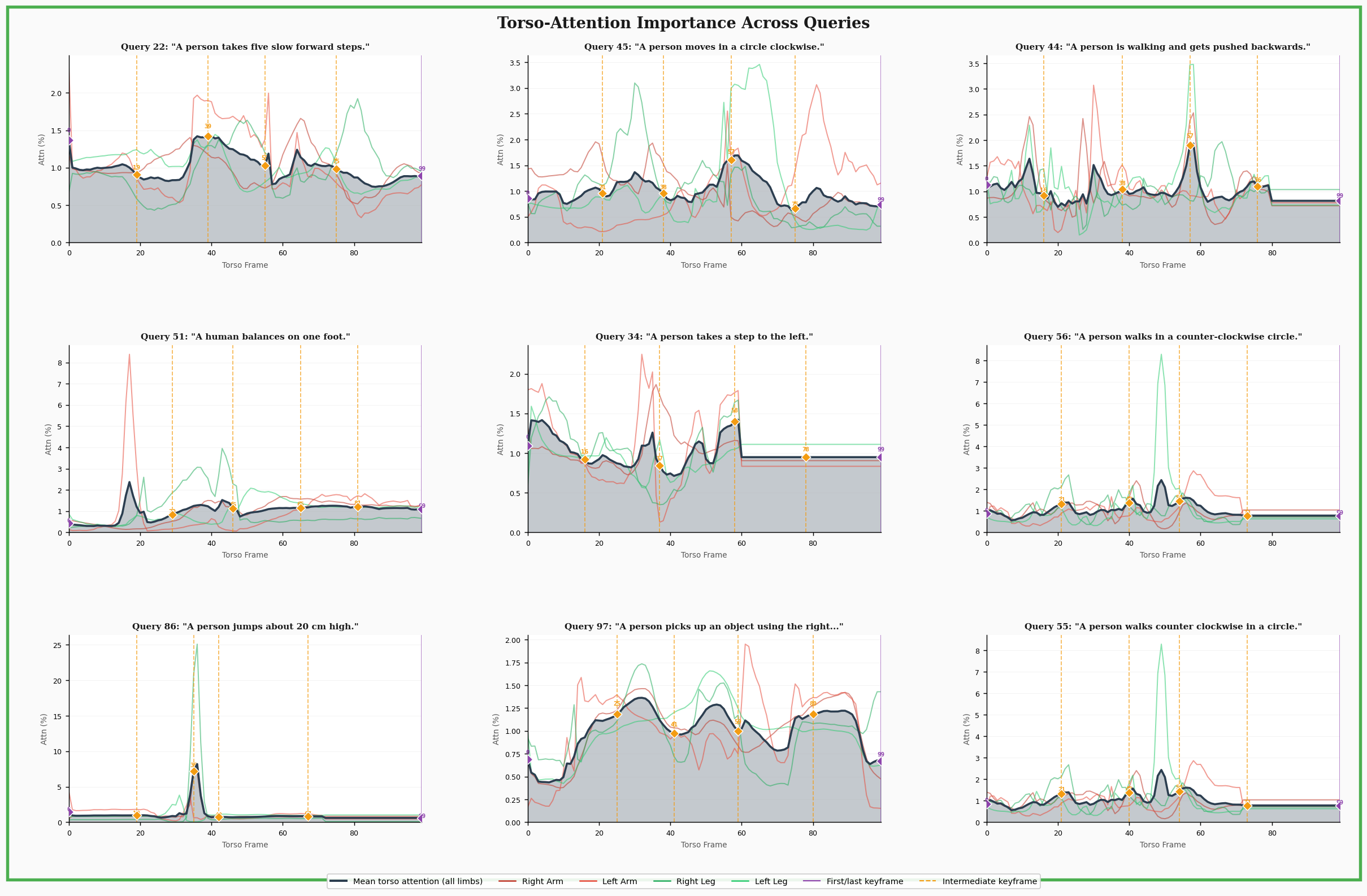}
    \caption{ Visualization of Torso-Centered Interaction learned by our proposed AttentionFuse module. In our 3T-hierarchical architecture, each limb branch (arms, legs) queries the torso branch to selectively retrieve temporally-relevant torso context, rather than encoding limb motion in isolation. For nine representative test queries, we plot the resulting torso-attention distribution: the x-axis is the torso frame index and the y-axis is the normalized attention weight (\%) each limb assigns to that frame. The bold navy curve is the mean attention across all four limbs (aggregate torso reliance); the thin red/green curves show per-limb (arm/leg) attention. Orange diamonds mark the frames of peak torso reliance, automatically extracted via cumulative-attention sampling.}
    \label{fig:TCI2}
\end{figure*}
The selected queries span simple upper-body actions (e.g., ``a person rubs their hands together''), sequential motions involving posture transitions (e.g., ``kneeling'', ``crawling'', and ``standing''), and complex compositional actions combining upper- and lower-body movements (e.g., ``stepping'' and ``kicking'', or repeated ``throwing'' actions). These examples highlight how textual verbs and action phrases are grounded in corresponding body-part movements, with the torso acting as a central coordinating component.

The visualization results of the skeleton representation and full-body SMPL demonstration, presented in Figures~5 and~6, demonstrate that HUMAN-TCI effectively captures both local and global motion patterns. For simple queries, the model retrieves precise fine-grained movements aligned with the described actions. For longer and compositional descriptions, it preserves temporal ordering and inter-part dependencies, correctly associating multiple verbs with sequential and overlapping motion segments. Notably, torso-driven actions such as crawling, kicking, and throwing are accurately retrieved, reflecting the model’s ability to capture how torso dynamics influence and regulate upper- and lower-body movements.

To illustrate torso-centered interaction at the level of an individual
query, we examine the query \textit{``A person takes five slow forward
steps''} (Figure~\ref{fig:TCI}). Each limb attends
non-uniformly to the torso sequence, with peak attention occurring at
frame~65 for the Right Arm ($1.66\%$), frame~0 for the Left Arm
($2.38\%$), frame~81 for the Right Leg ($1.92\%$), and frame~41 for the
Left Leg ($1.44\%$), the four peaks are visually distinct and
largely non-overlapping (Figure~\ref{fig:TCI}, top row and
heatmap), indicating each limb is anchoring to a different torso
configuration rather than converging on a single dominant frame. The
attention entropy for all four limbs is similar ($H \approx 4.5$--$4.6$),
suggesting a comparably diffuse-but-peaked distribution across limbs
for this query, rather than one limb dominating the torso context while
others attend near-uniformly. The cumulative attention curve
(Figure~\ref{fig:TCI}, bottom-right) shows that $50\%$ of
total torso-attention mass is accumulated by frame~47 ($F_{50}$) and
$90\%$ by frame~88 ($F_{90}$), indicating the model concentrates the
majority of its torso reliance within roughly the first half of the
sequence, consistent with the ``five slow forward steps''
description, where the gait pattern is established early and
subsequent frames largely reinforce rather than introduce new torso
context. Notably, the Right Leg and Left Leg attention peaks (frames~81
and~41) are separated by roughly 40 frames, plausibly corresponding to
the alternating left--right foot-strike pattern inherent to walking,
while the Left Arm's peak at frame~0 reflects the model anchoring to
the initial standing posture before forward motion begins. This
per-query breakdown provides direct, frame-level evidence that our
proposed torso-centered interaction mechanism dynamically and
selectively routes each limb's representation through kinematically
distinct torso states, rather than applying a fixed or redundant
attention pattern.

A central design principle of our 3T-hierarchical architecture is torso-centered interaction: rather than encoding each limb's motion independently, every limb branch (arms, legs) queries the torso branch via AttentionFuse, using the torso's temporal sequence as the source of Key and Value context. This is motivated by the observation that torso pose and orientation act as the anchor around which limb articulation is organized, the torso provides the frame of reference (e.g., facing direction, balance, weight shift) that gives limb movement its meaning. To verify that the model actually exploits this design rather than treating it as a redundant pathway, we visualize the learned torso-attention weights in Figure~\ref{fig:TCI2}. Across all examined queries, attention is sharply non-uniform: each limb selectively concentrates its reliance on a small number of torso frames rather than distributing attention uniformly across the sequence, which it is free to do since torso attention is not explicitly supervised. This confirms that the model has learned to treat the torso as an active, frame-selective reference signal, consistent with our architectural hypothesis, rather than as a static or uninformative context. Moreover, the location of peak torso attention aligns with visually salient torso configurations (e.g., a shift in stance mid-stride, or the torso's compressed pose at the apex of a jump), indicating the model anchors limb representations to kinematically meaningful torso states, not arbitrary frames. This provides direct provides qualitative evidence for the torso-centered interaction our architecture is designed around.

Overall, these qualitative results confirm that HUMAN-TCI robustly aligns textual semantics with human motion by explicitly modeling torso-centered interactions. This enables effective understanding of action verbs, compositional structures, and coordinated motion patterns, leading to semantically consistent and physically plausible retrieval outcomes.

\section{Conclusion and Future Work}

In this work, we addressed the challenging task of text-to-motion retrieval, particularly focusing on the limitations of existing approaches in handling compositional language and complex body-part dependencies. While prior methods often rely on holistic or independently modeled representations, they fail to explicitly capture the hierarchical and interactive nature of human motion, especially the influence of torso dynamics on coordinated body movements. To overcome these limitations, we proposed \textbf{HUMAN-TCI}, a Hierarchical Multi-Stream Motion-Aware Network that explicitly models upper-body, lower-body, and torso interactions through a structured multi-stream architecture. By introducing a torso-aware attention mechanism, our approach enables fine-grained interaction between body parts, allowing the model to better capture complex motion semantics and dependencies. This design makes our framework effective for both simple action descriptions and long, compositional sentences involving sequential or overlapping actions. Extensive experiments on the KIT Motion-Language Dataset and HumanML3D demonstrate that HUMAN-TCI consistently achieves superior performance compared to existing methods across multiple evaluation metrics, including Recall@K, Median Rank, and Mean Rank. The results confirm that our method improves retrieval accuracy, enhances semantic alignment and interpretability, and maintains computational efficiency suitable for large-scale applications.

Despite these promising results, several directions remain for future work. We aim to enhance hierarchical semantic alignment through multi-level reasoning to better capture long-range dependencies in complex textual descriptions. Additionally, leveraging cross-modal pretraining and multimodal foundation models may improve generalization and robustness. Incorporating additional modalities such as video or depth could further enrich motion understanding, while exploring finer skeletal representations and motion decomposition may enable more precise and interpretable retrieval.

\bibliographystyle{elsarticle-num}


\bibliography{cas-refs}
\end{document}